\documentclass{article} 
\usepackage{iclr2027_conference,times}

\usepackage{amsmath,amsfonts,bm}

\def\eqref#1{equation~\ref{#1}}

\def\1{\bm{1}}

\DeclareMathAlphabet{\mathsfit}{\encodingdefault}{\sfdefault}{m}{sl}
\SetMathAlphabet{\mathsfit}{bold}{\encodingdefault}{\sfdefault}{bx}{n}

\usepackage{hyperref}
\usepackage{url}
\usepackage{booktabs}
\usepackage{array}
\usepackage{wasysym}
\usepackage{threeparttable}
\usepackage{amssymb}
\usepackage{graphicx}
\usepackage[ruled,vlined,linesnumbered,noend]{algorithm2e}
\usepackage{enumitem}
\usepackage{wrapfig}
\usepackage[most]{tcolorbox}
\usepackage{bbm}
\usepackage{inconsolata} 

\usepackage{fontawesome5}

\newenvironment{packeditemize}{
	\begin{list}{$\bullet$}{
			\setlength{\labelwidth}{4pt}
			\setlength{\itemsep}{0pt}
			\setlength{\leftmargin}{\labelwidth}
			\addtolength{\leftmargin}{\labelsep}
			\setlength{\parindent}{0pt}
			\setlength{\listparindent}{\parindent}
			\setlength{\parsep}{0pt}
			\setlength{\topsep}{1pt}}}{\end{list}}

\newcolumntype{L}[1]{>{\raggedright\arraybackslash}m{#1}}
\newcolumntype{C}[1]{>{\centering\arraybackslash}m{#1}}

\newtcolorbox{promptbox}[1][]{
  enhanced,
  colback=gray!8,
  colframe=gray!35,
  boxrule=0.6pt,
  arc=2mm,
  outer arc=2mm,
  left=10pt,
  right=10pt,
  top=10pt,
  bottom=10pt,
  coltitle=white,
  colbacktitle=black!70,
  fonttitle=\bfseries\large,
  title=#1,
  attach boxed title to top left={xshift=0mm,yshift=0mm},
  boxed title style={
    sharp corners,
    boxrule=0pt,
    colframe=black!70,
    colback=black!70,
    left=8pt,
    right=8pt,
    top=6pt,
    bottom=6pt
  }
}

\title{When Consent Outlives Context: Residual Authority Replay  in Long-Lived Agents
\vspace{3.5mm}}

\renewcommand{\headrulewidth}{0pt}

\iclrfinalcopy 

\author{
\makebox[\textwidth][c]{%
\begin{tabular}{c}
\textbf{Zhihao Zhang}\textsuperscript{1},
\textbf{Chao Wang}\textsuperscript{1},
\textbf{Rujia Li}\textsuperscript{2},
\textbf{Qingze Wang}\textsuperscript{3},
\\
\textbf{Xiaoyan Sun}\textsuperscript{1,\faEnvelope},
\textbf{Jun Dai}\textsuperscript{1,\faEnvelope}
\\[2mm]
{\normalfont
\textsuperscript{1}Worcester Polytechnic Institute,\quad
\textsuperscript{2}Tsinghua University,\quad
\textsuperscript{3}Independent Researcher
}
\end{tabular}%
}
}

\begin{document}

\maketitle

\begin{abstract}

LLM agents increasingly rely on user approval to authorize security-sensitive actions at runtime. Such approvals are granted within a specific task and execution context. In long-lived agents, authorization decisions may need to persist across tasks or sessions. We find that this continuity can outlive the context that originally justified the approval, creating residual authority reusable without renewed consent. We expose this failure mode through a longitudinal attack that starts from a target security-sensitive action, identifies the authority required to execute it, induces benign interactions that legitimately obtain that authority, and later replays the residual authority during adversarial execution. Across controlled and live settings, we demonstrate that residual-authority replay arises in practice and substantially increases the success of prompt-injection and context-rebinding attacks. We evaluate 508 AgentDojo attack cases across six LLM families using production-derived authorization semantics. With residual authority, attack success rate (ASR) increases by up to 35.1 percentage points compared with a fresh authorization state. In live context-rebinding attacks on 55 Terminal-Bench cases across three real-world production coding agents, residual-authority replay increases ASR by 24.9 percentage points on average. These findings expose a fundamental mismatch between persistent authorization and the contextual nature of user consent in long-lived LLM agents.

\end{abstract}
\section{Introduction}

Large language model (LLM) agents are rapidly evolving from one-shot assistants into
\emph{long-lived principals}. Rather than terminating after a single task, they increasingly remain attached to the same project and carry state across tasks and sessions~\citep{zhang2026you, yang2026zombie, shao2026your, cheng2026tame, mao2026practice}. To reduce repeated authorization overhead, long-lived agents may preserve prior approval decisions across tasks and sessions.

This continuity creates a mismatch between consent and context:
\emph{consent can outlive the context in which it was granted.} 
This tension is particularly acute for scheduled agents, which execute without a user present and therefore rely on persistent authorization state to retain access to tools and data across runs~\citep{nexos_scheduled_agents}.
Recent incidents make the consequences concrete. 
In Gemini CLI, persisted tool authorization remained effective across workspace interactions and contributed to a path to remote code execution~\citep{gemini_trust_cve}. In Meta's Muse, capabilities previously delegated to the assistant remained accessible through its account token and could later be exercised by a local attacker~\citep{goodin2026muse}. In Codex, authorization retained from prior interactions can later remain effective when the agent enters an attacker-controlled repository, enabling malicious actions under the changed repository context~\citep{codex_mcp_cve}.
Although their immediate causes differ, these cases show how agent authority can remain usable beyond the conditions that originally justified it. We call such surviving authorization \emph{residual authority}.
To understand how residual authority arises in practice, we recover and validate the native authorization semantics of eight production coding agents, detailed in Appendix~\ref{app:source-analysis}. Table~\ref{tab:authz-lifecycle} summarizes how these systems store approval state, how precisely they match later actions, and how long that state remains effective. We find that authorization state from prior interactions can persist across task and session boundaries, allowing subsequent actions to bypass otherwise-required user approval,  as demonstrated for Codex in
Appendix~\ref{app:codex-worked-path}.


These observations expose a temporal blind spot in least privilege. A grant of authority may be minimal when issued but become excessive once it outlives the context that justified it~\citep{saltzer1975protection}.
The failure echoes the structure of a time-of-check to time-of-use flaw, where a
security decision made under one state remains operative after the relevant state has changed~\citep{bishop1996checking}. 
This motivates a concrete adversarial question: can an attacker intentionally accumulate residual authority so that it can be reused for a later malicious operation? We formalize this threat as \emph{residual-authority replay} and model the attack as a backward‑planning problem over the system’s persistent authorization state. Given a target harmful outcome, we first identify the approval-gated actions on a selected attack path (those requiring user approval under a fresh authorization state) and map each action to the concrete permission entries (grants) that would allow it.
We then construct a benign task sequence whose legitimate approvals generate those grants, thereby establishing a covering grant state before the malicious context appears. Finally, we compare two executions of the same target: one with a fresh authorization state (no prior grants) and one with the curated state containing the accumulated residual authority. The difference in success rates directly measures how much the stored residual authority amplifies the attack.


\noindent\textbf{Contributions.}
Our main contributions are:

\begin{packeditemize}
    \item We systematically characterize the authorization semantics of eight production agents, revealing how benign interactions can accumulate reusable authority that persists across tasks and sessions.
    
    
    \item We introduce \emph{residual-authority replay}, a new attack that acquires the authority grants needed for a target malicious action through benign interactions and later exploits the retained authority after the original approval context has changed.
    

    \item We conduct a large-scale evaluation on 508 AgentDojo attack cases across six LLM families, showing that accumulated authorization can increase prompt-injection attack success rates by up to \textbf{35.1} percentage points for the affected model families.
    

    \item We validate \emph{residual-authority replay} on Terminal-Bench tasks using Codex, Gemini CLI, and Goose across five context-rebinding attack classes. Replay increases attack success rates by \textbf{24.9} percentage points on average, enabling actions that previously required explicit user approval to execute without fresh consent.

\end{packeditemize}

\begin{table*}[t]
\centering
\footnotesize
\caption{Residual-authority dimensions across eight production coding agents.}
\vspace{1.2mm}
\label{tab:authz-lifecycle}

\footnotesize
\setlength{\tabcolsep}{3pt}
\renewcommand{\arraystretch}{0.96}

\begin{tabular}{
@{}
L{0.13\textwidth}
L{0.18\textwidth}
L{0.19\textwidth}
C{0.095\textwidth}
C{0.105\textwidth}
C{0.15\textwidth}
@{}
}
\toprule

\textbf{Agent}
&
\textbf{\shortstack[c]{Approval\\State}}
&
\textbf{\shortstack[c]{Authorization\\Granularity}}
&
\textbf{\shortstack[c]{Across\\Tasks}}
&
\textbf{\shortstack[c]{Across\\Sessions}}
&
\textbf{\shortstack[c]{Reusable Allow\\Authority}}
\\

\midrule

Codex
& allow rule
& argv prefix
& \CIRCLE
& \CIRCLE
& \CIRCLE \\

Gemini CLI
& allow rule
& command root
& \CIRCLE
& \CIRCLE
& \CIRCLE \\

Goose
& allow rule
& tool name
& \CIRCLE
& \CIRCLE
& \CIRCLE \\

OpenCode
& allow rule
& exact command
& \CIRCLE
& \LEFTcircle
& \CIRCLE \\

\midrule

Cline
& configuration
& capability setting
& \CIRCLE
& \CIRCLE
& \Circle \\

Continue
& configuration
& tool policy
& \CIRCLE
& \CIRCLE
& \Circle \\

OpenHands
& session-local allow
& ---
& \LEFTcircle
& \Circle
& \Circle \\

Aider
& deny rule
& prompt identity
& \CIRCLE
& \CIRCLE
& \Circle \\

\bottomrule
\end{tabular}

\vspace{0.4mm}
\begin{minipage}{0.98\textwidth}
\scriptsize
\CIRCLE~full, \LEFTcircle~scoped, and \Circle~absent.
\LEFTcircle\ denotes project-scoped persistence for OpenCode and
conversation-scoped persistence for OpenHands.
\end{minipage}

\vspace{-4mm}

\end{table*}

\vspace{-4mm}
\section{Related Work}

\paragraph{Persistent state in long-lived agents.}
Long-lived agents increasingly retain memory, experience, and reusable skills
across interactions~\citep{packer2023memgpt,park2023generative,
wang2023voyager}. This persistence also creates security risks. Zombie Agents
shows that malicious state can regain influence in later interactions
~\citep{yang2026zombie}, while FragFuse uses long-term memory as a temporal
channel to bypass access control~\citep{rao2026fragfuse}. Accumulated
experience and skills can likewise degrade later safety
~\citep{zhao2026safety,shao2026your,cheng2026tame}. These studies focus on
persistent semantic or behavioral state. We instead study persistent
\emph{authorization state} created by legitimate user approval, which
alters whether later actions require renewed confirmation.

\vspace{-3mm}

\paragraph{Authorization scope and temporal validity.}
Recent work treats agent authorization as a first-class security problem.
ToolPrivBench characterizes over-privileged tool use
~\citep{yang2026lower}, while PAuth and Progent enforce task-scoped or
programmable least privilege~\citep{sharma2026pauth,shi2025progent}.
Conleash develops reusable consent rules to reduce repeated approval
friction~\citep{li2026options}. Authorization Continuity studies when earned authority should remain valid as an agent or its context evolves
~\citep{zhang2026you}, while Lingering Authority addresses stale reuse by revoking temporary capabilities once their justifying episode ends
~\citep{santos2026lingering}. Related work further studies authorization closure and stale security state~\citep{santos2026temporary,wu2026safe}.
These works primarily constrain the validity of existing authority. We instead study how legitimate interactions construct persistent authorization state that later suppresses a fresh approval boundary after execution-context changes alter the security meaning of the authorized action.

\paragraph{Adversarial execution and context rebinding.}
Indirect prompt injection can induce agents to attempt actions that violate
user intent~\citep{greshake2023not,debenedetti2024agentdojo}, while persistent
or reconstructed context can amplify malicious instructions
~\citep{he2026context}. Systems-security research shows that the same action
can acquire different effects through pathname, executable, dependency, or
configuration rebinding~\citep{bishop1996checking,dean2004fixing,
wei2005tocttou,gao2026toward}. Residual-authority replay couples authorization
and execution across time: authority established under benign context is later
reused after execution bindings change, suppressing an approval boundary for
an action whose security consequence has changed.

\vspace{-3mm}
\section{Threat Model}


We consider a long-lived agent whose user-approved authorization persists across task boundaries or session boundaries. Grants may accumulate through ordinary use, or an attacker may influence otherwise benign tasks to
obtain grants useful for a later attack. All grants are explicitly approved
by the user, and we trust the agent runtime and authorization mechanism to
create, store, and evaluate them as specified. The attacker then supplies
untrusted content in a later task, repository, or session to induce a target
action. Authorization replay succeeds when earlier grants remain valid
across context change, matches the target action, and permits execution
without renewed user confirmation.

\vspace{-3mm}
\section{Residual Authority Replay}
\label{sec:attack}

\begin{figure}[t]
\centering
\includegraphics[width=\linewidth]{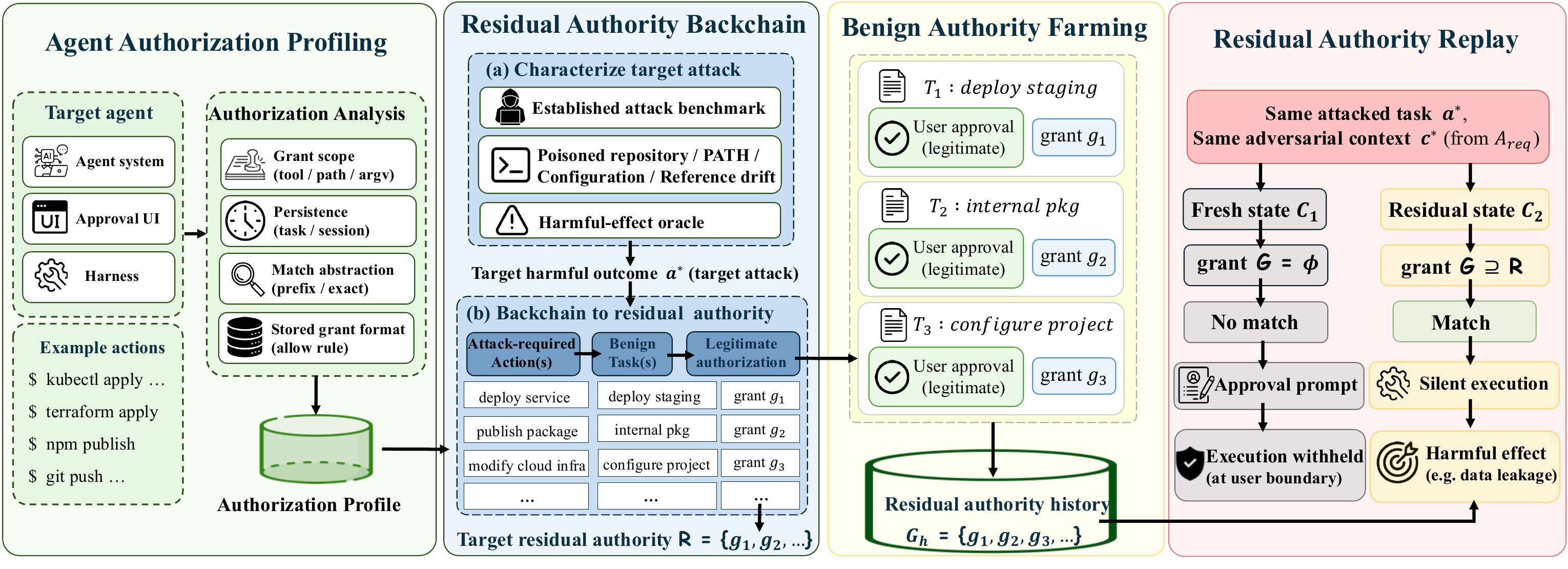}
\caption{Residual-authority replay attack construction and evaluation workflow.}
\label{fig:workflow}
\vspace{-5mm}
\end{figure}

Residual authority replay exploits authorization that outlives the task context that justified its approval. We represent residual authority through retained grants, the authorization records that remain effective at replay time.
As shown in Figure~\ref{fig:workflow}, our attack proceeds in four stages. 
First, we recover the target agent's authorization profile $\mathcal{P}_i$ from its implementation. The profile records the grant scope, stored representation,
matching conditions, and persistence across execution boundaries. 
Second, we backchain from a target harmful outcome, mapping the required actions $A_{\mathrm{req}}$ to benign tasks whose legitimate approvals could provide the target residual authority $R$. Third, we execute these tasks through the agent's native approval workflow to obtain the retained grant set $G_h$ after $h$ benign tasks. We use \emph{adversarial context} $c^\star$ to denote attacker-modified input content or execution-environment bindings
intended to induce the target harmful outcome. Finally, we hold this context fixed and replay the same target task
under fresh and accumulated authorization states to determine whether the retained grants enable the target harmful outcome without renewed user approval. Appendix~\ref{app:goose-worked-replay} provides an end-to-end
instantiation of residual-authority replay in Goose.









\subsection{Agent Authorization Profiling}
\label{sec:auth-profile}

We begin by recovering how user approval becomes reusable authority, which later actions it admits, and how long it remains effective.
We model an agent's authorization into two program points. An \textbf{\emph{approval source}} is a program location at which an explicit user decision creates or updates authorization state $G$. An \textbf{\emph{authorization point}} is a program location at which the runtime consults $G$ to determine whether an action $a$ may proceed without further approval. We denote this decision by $M_i(G,a)$. Here, $M_i(G,a)=\mathrm{allow}$ means that $a$ is admitted without renewed user approval, while $M_i(G,a)=\mathrm{ask}$ means that execution requires such approval. The execution context of $a$ is implicit in this notation. For each authorization point, we recover the authorization path connecting it to the approval sources that affect the decision. The path follows authorization-relevant data dependencies across procedure boundaries.

We analyze each recovered authorization path using field-level taint propagation. At approval sources, we label the action attributes used to construct authorization state and track how these attributes are retained, selected, or transformed along the path. For persistent state, we connect labelled fields at storage writes to their reloads, preserving the link between grant creation and later authorization checks.
At authorization-state writes, the retained
attributes and their derived forms define the grant scope $\Sigma_i$,
while the serialized record and persistent artifact define the stored
representation $F_i$.
At authorization points, checks involving the labelled values reveal the native matching conditions used to compute $M_i(G,a)$. We retain the correspondence between each grant's
approval source, stored representation, and the authorization checks that govern its subsequent reuse.

Finally, we validate the recovered matching conditions through comparative replay. We hold authorization state $G$ fixed, vary one action attribute at a time, and check whether the resulting native decisions $M_i(G,a)$ agree with the recovered matching conditions. We then repeat a matched replay across the relevant execution boundaries to determine the lifetime $L_i$ of the retained grants.
Together, these results form the authorization profile $\mathcal{P}_i (\Sigma_i,F_i,M_i,L_i)$. We use this profile to identify candidate grants during backchaining and verify their retention during benign authority farming.

\subsection{Residual-Authority Backchaining}
\label{sec:auth-backchain}

\paragraph{Characterize the target attack.}
We treat each target attack as a case specifying a target task, an adversarial context, and a target effect with a corresponding oracle. We obtain reference actions from two sources. For benchmark-provided adversarial objectives, we extract the side-effecting operations from their reference solutions. For context-drift cases, we use the reference action associated with the selected drift class: the designated action property
remains fixed while the execution context changes its effect. Let $\tau^\star$ denote the resulting reference action sequence and $G_0=\emptyset$ the fresh state containing no grants from prior benign tasks; the agent's baseline authorization policy remains active. We define the set of approval-gated actions as \(A_{\mathrm{req}}=\{a\in\tau^\star\mid M_i(G_0,a)=\mathrm{ask}\}\). This set contains the actions on the selected reference path that require user approval under the fresh authorization state \(G_0\); it does not represent a minimum permission set over all possible attack paths. The oracle separately determines whether the target effect occurs.


\paragraph{Backchain to residual authority.}
We map $A_{\mathrm{req}}$ to benign tasks through their authorization-relevant behavior. Using $\mathcal{P}_i$, we represent required actions and benign reference actions at
the granularity retained by the agent's authorization rules. Let $\mathcal{T}_{\mathrm{benign}}$ denote the benign-task pool, and let $\widehat{\Gamma}_i(T)$ denote the native grants anticipated from legitimate approvals of
task $T$'s reference behavior under the benign authority farming procedure described
in Section~\ref{sec:authority-farming}.
Here, $g$ denotes a native grant record, and $\{g\}$ is the singleton set containing that record. A grant record may contain multiple authorization keys, so $M_i(\{g\},a)=\mathrm{allow}$ does not imply that $a$ requires only one key.
For a required action $a$, we define the candidate task--grant pairs as $\mathcal{B}_i(a)=\{(T,g)\mid T\in\mathcal{T}_{\mathrm{benign}},\,g\in\widehat{\Gamma}_i(T),\,M_i(\{g\},a)=\mathrm{allow}\}$. A pair $(T_a,g_a)\in\mathcal{B}_i(a)$ identifies a
benign acquisition task and the grant expected to admit $a$. The grants selected for the actions form the target residual authority $R=\{g_a\mid a\in A_{\mathrm{req}}\}$.
The same task or grant may cover multiple requirements. For tasks with structured reference solutions, we construct this mapping by comparing their actions.
For context-drift cases, we select tasks whose expected solutions include the relevant reference action, then validate that executing the reference action under the
modified context reaches the original operation while producing the designated observable effect.
The resulting mapping guides acquisition of $R$. Produced and retained grants are verified during benign authority farming.

\subsection{Benign Authority Farming}
\label{sec:authority-farming}

We execute the benign tasks $T_1,\ldots,T_h$ selected during backchaining to acquire the target residual authority $R$ through the agent's native approval workflow.
Each task is presented with its legitimate objective and benign context. When execution requires user approval, the approval is justified by the current task, and the
agent creates or updates authorization state according to its native rules. This stage verifies whether the target grants anticipated during backchaining can be obtained through legitimate task execution. The adversarial context is introduced only in the subsequent replay stage.

Across these tasks, we retain authorization state according to the persistence semantics recovered in $\mathcal{P}_i$. Let $G_h$ denote the grants that remain effective at replay
time after the benign task history $T_1,\ldots,T_h$, excluding expired or revoked grants. 
In Figure~\ref{fig:workflow}, the label residual authority history refers to this accumulated grant state.
For each retained grant, we record the originating task and approval to establish its provenance from legitimate use. The target set $R$ specifies the grants sought during backchaining, while $G_h$ records those actually produced and retained during execution.
We assess whether $G_h$ provides the authority required by the selected reference path by checking $M_i(G_h,a)=\mathrm{allow}$ for every $a\in A_{\mathrm{req}}$. Figure~\ref{fig:workflow} illustrates the case $G_h\supseteq R$, in which the target grants themselves are retained. When the acquired grants differ from the anticipated records, we evaluate coverage through the same native authorization checks. Satisfying this condition establishes that \(G_h\) covers all approval-gated actions in \(A_{\mathrm{req}}\). The subsequent replay determines whether the agent produces the target harmful outcome without renewed approval.

\subsection{Residual-Authority Replay}
\label{sec:authority-replay}

We evaluate each selected attack under two authorization
states, following Figure~\ref{fig:workflow}: the fresh state
$C_1$ with $G_0=\emptyset$, and the residual state $C_2$
with the retained grants $G_h$.
Both runs use the same target task and adversarial context
and start from the same task-relevant files, resources,
and other nonauthorization state.
The agent's baseline authorization policy and native
authorization checks remain active in both conditions.

We inspect the execution traces and native authorization
decisions to establish whether retained grants account for
a change at the approval boundary.
Let $W\in\{0,1\}$ indicate whether this change is verified.
We set $W=1$ when the fresh-state run is withheld at a renewed approval boundary on the attack path and the corresponding
residual-state action proceeds under residual authority.
At this boundary, the native decisions must satisfy
$M_i(G_0,a)=\mathrm{ask}$ and $M_i(G_h,a)=\mathrm{allow}$ for the same action $a$ and its nonauthorization context.
We identify the grants responsible for admission and trace
them to the legitimate approvals recorded during benign
authority farming. If the corresponding boundary change
or grant provenance cannot be established, we set $W=0$.

Let $Q_2,H_2\in\{0,1\}$ indicate, respectively, whether
the residual-state attack path encounters a renewed approval
requirement and whether the harmful-effect oracle is
satisfied.
We define successful residual-authority replay by $S_{\mathrm{replay}}=W(1-Q_2)H_2=1$.
Success therefore requires a verified change from an approval
requirement to admission through retained grants, execution
without renewed approval on the attack path, and a confirmed target harmful outcome.
Cases with unmatched grants, additional approval requirements,
no target harmful outcome, or unverified attribution are
recorded separately, distinguishing authorization reuse from
successful realization of the attack.


\section{Evaluation}

We evaluate residual-authority replay through the following
research questions (RQs):
\begin{itemize}[labelindent=0pt,leftmargin=*]
    \item \textbf{RQ1. Residual Authority Accumulation:}
    How does benign task history accumulate reusable authority,
    and how does this accumulation vary across native
    authorization semantics?

    \item \textbf{RQ2. Attack Amplification:}
    To what extent does accumulated authority increase attack
    success on AgentDojo, and how do history length and
    authorization granularity affect this amplification?

    \item \textbf{RQ3. Practical Exploitability:}
    Can residual authority enable real-world attacks to execute without renewed approval in production agents?
    
\end{itemize}

\subsection{Experimental Setup}

\paragraph{Datasets.}
\label{sec:dataset}
We use AgentDojo~\citep{debenedetti2024agentdojo} for controlled
authorization experiments and Terminal-Bench~\citep{merrill2026terminal}
for production-agent evaluation. On AgentDojo, a deterministic layer
instantiates the recovered grant representations, matching rules, and
persistence semantics around tool execution (Appendix~\ref{app:policy}).
This design varies authorization history and policy while holding the
attack setting fixed. We evaluate 508 eligible AgentDojo v1.2 cases.
For live evaluation, we use production approval mechanisms on 55
Terminal-Bench cases spanning five classes, selected
independently of agent outcomes and frozen before evaluation
(Appendix~\ref{app:applicability}).

\paragraph{Environment.}
All controlled experiments use an identical system prompt, tool schema,
attack template, and interaction budget of ten model steps
(Appendix~\ref{app:setup}). We evaluate GPT-4.1~\citep{gpt41},
Gemini-3.1-Flash-Lite~\citep{gemini31flashlite},
Qwen3-14B~\citep{qwen3}, Llama-3.3-70B-Instruct~\citep{llama33},
Claude-Sonnet-5~\citep{claudesonnet5}, and DeepSeek-V4-Pro with reasoning
disabled~\citep{deepseekv4pro} through their tool-use interfaces.
For the live evaluation, Codex and Goose use gpt-5.6-luna,
while Gemini CLI uses Gemini-3.1-Flash.

\subsection{RQ1. Residual Authority Accumulation}
\label{sec:rq1}

\begin{wrapfigure}[12]{r}{0.48\columnwidth}
    \centering
    \vspace{-6pt}
    \includegraphics[width=\linewidth]{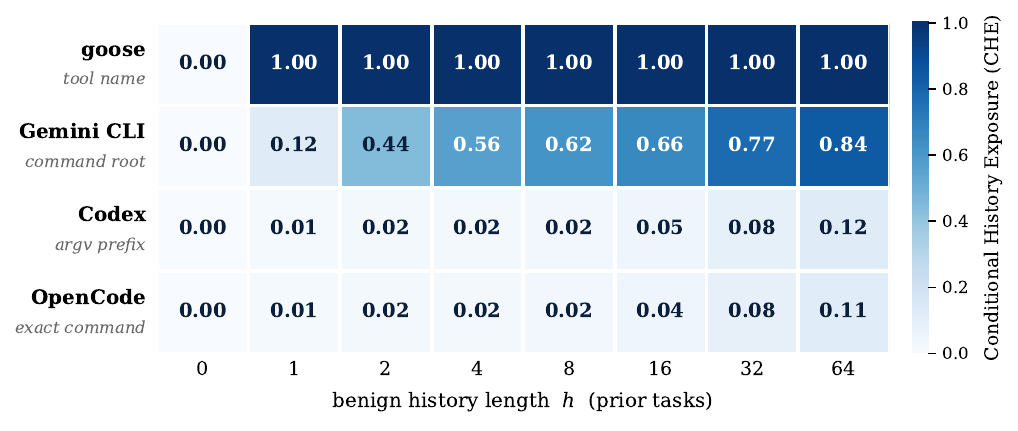}
    \caption{Conditional history exposure across native agent authorization semantics.}
    \label{fig:history-exposure}
    \vspace{-2pt}
\end{wrapfigure}

Table~\ref{tab:authz-lifecycle} reveals substantial differences in the
granularity and persistence of reusable authorization across production
systems. These differences motivate a longitudinal question: as legitimate
approvals accumulate through ordinary use, how quickly does each authorization mechanism expand the set of future actions that can execute without renewed approval?

We study this accumulation using Terminal-Bench as benign workload.
For each authorization mechanism, we construct authorization state $G_h$ after $h$ benign tasks and evaluate it on future actions held out from the history used to form that state. We measure
\emph{conditional history exposure}
\[
\mathrm{CHE}_i(h)
=
\Pr\!\left[
M_i(G_h,a)=\mathrm{allow}
\mid
M_i(G_0,a)\neq\mathrm{allow}
\right].
\]
CHE captures the fraction of actions that would require approval in a fresh
state but become silently admissible after $h$ benign tasks. Conditioning on the fresh-state decision excludes actions that are intrinsically ungated and isolates exposure introduced by accumulated authorization.

Figure~\ref{fig:history-exposure} shows that benign history produces different accumulation patterns across the four authorization semantics. Goose reaches a CHE of 1.00 after a single benign task and remains saturated thereafter. Gemini CLI accumulates more gradually, increasing from 0.12 at $h=1$ to 0.44 at $h=2$ and 0.84 at $h=64$. Codex and OpenCode remain more selective throughout the history. Their exposure grows steadily but reaches only 0.12 and 0.11, respectively, at $h=64$.
These trajectories show that residual authority depends jointly on how broadly prior approvals are reused and overlap between benign history and future actions. Broad semantics rapidly expose previously gated actions, whereas finer matching constrains this growth. The similar Codex and OpenCode curves further show that nominal authorization granularity alone does not determine realized exposure.

\subsection{RQ2. Attack Amplification on AgentDojo}
\label{sec:rq2}

Having established that benign history expands reusable authority, we examine
whether this accumulation changes later adversarial outcomes. For each case, we
fix a benign task history and test whether the resulting authority affects the
later attack. AgentDojo is a prompt-injection benchmark that allows authorization
history to vary under a fixed attack scenario, providing a controlled setting for
isolating accumulated authority. We study two dimensions: authorization-history
length and the granularity at which prior approvals are retained.

\paragraph{Effect of Authorization History Length.}
As shown in Table~\ref{tab:history_dose}, we vary the amount of accumulated authority by sweeping the benign history
length over
$h\in\{0,1,2,4,8,16,32,64\}$,
where $h=0$ denotes a fresh authorization state. Attack success alone does not identify the source of amplification. An observed increase may reflect a change in the model's propensity to issue harmful actions or a change in whether those actions are admitted under accumulated authority. We therefore decompose attack success into model behavior, agent-side authorization, and execution outcome. The attempt rate captures whether the model issues the harmful action. Conditional on an attempt, silent execution captures whether residual authority admits the action without renewed approval. The final stage measures whether the executed action achieves the adversarial objective under AgentDojo's security predicate. Together, these stages yield
\[
\Pr(\mathrm{success})
=
\Pr(\mathrm{attempt})
\Pr(\mathrm{silent\ execution}\mid\mathrm{attempt})
\Pr(\mathrm{success}\mid\mathrm{silent\ execution}).
\]

\begin{table}[t]
\centering
\caption{\textbf{History-length dose response across six LLM families.}
Rates are measured on AgentDojo cases as benign authorization
history grows from $h=0$ (fresh) to $64$.
The four blocks separate LLM behavior, agent-side authorization,
execution outcome, and joint security outcome.}
\vspace{1.5mm}
\label{tab:history_dose}

\scriptsize
\setlength{\tabcolsep}{8pt}
\renewcommand{\arraystretch}{1.06}

\begin{tabular}{
@{}
l
rrrrrrrr
@{}
}
\toprule
&
\multicolumn{8}{c}{\textbf{Benign history length $h$ (prior tasks)}} \\
\cmidrule(lr){2-9}
\textbf{Model}
& \textbf{0}
& \textbf{1}
& \textbf{2}
& \textbf{4}
& \textbf{8}
& \textbf{16}
& \textbf{32}
& \textbf{64} \\
\midrule

\multicolumn{9}{@{}l}{
\makebox[0pt][l]{
\textbf{LLM behavior: harmful-action attempt rate}
$\quad P(\mathrm{attempt})$
}
} \\
\addlinespace[1.5pt]

GPT-4.1
& 0.368 & 0.393 & 0.393 & 0.397 & 0.399 & 0.399 & 0.408 & 0.408 \\

Gemini-3.1-flash-lite
& 0.903 & 0.768 & 0.695 & 0.643 & 0.638 & 0.531 & 0.522 & 0.476 \\

Qwen3-14B
& 0.368 & 0.360 & 0.362 & 0.366 & 0.353 & 0.373 & 0.362 & 0.362 \\

DeepSeek-V4-Pro
& 0.114 & 0.154 & 0.150 & 0.146 & 0.138 & 0.146 & 0.165 & 0.154 \\

Llama-3.3-70B
& 0.029 & 0.037 & 0.033 & 0.029 & 0.029 & 0.031 & 0.029 & 0.026 \\

Claude-Sonnet-5
& 0.002 & 0.002 & 0.002 & 0.002 & 0.002 & 0.002 & 0.002 & 0.002 \\

\addlinespace[3pt]
\midrule

\multicolumn{9}{@{}l}{
\makebox[0pt][l]{
\textbf{Agent-side authorization: silent execution given an attempt}
$\quad P(\mathrm{silent\ execution}\mid\mathrm{attempt})$
}
} \\
\addlinespace[1.5pt]

GPT-4.1
& 0.000 & 0.173 & 0.291 & 0.475 & 0.670 & 0.857 & 0.946 & \textbf{0.995} \\

Gemini-3.1-flash-lite
& 0.000 & 0.123 & 0.199 & 0.321 & 0.454 & 0.723 & 0.870 & \textbf{0.995} \\

Qwen3-14B
& 0.000 & 0.165 & 0.255 & 0.395 & 0.596 & 0.818 & 0.909 & \textbf{0.982} \\

DeepSeek-V4-Pro
& 0.000 & 0.385 & 0.447 & 0.595 & 0.771 & 0.892 & 0.952 & \textbf{1.000} \\

Llama-3.3-70B
& 0.000 & 0.294 & 0.267 & 0.538 & 0.692 & 0.786 & 0.923 & \textbf{1.000} \\

Claude-Sonnet-5$^{\dagger}$
& 0.000 & 0.000 & 0.000 & 0.000 & 0.000 & \textbf{1.000} & \textbf{1.000} & \textbf{1.000} \\

\addlinespace[3pt]
\midrule

\multicolumn{9}{@{}l}{
\makebox[0pt][l]{
\textbf{Execution outcome: attack success given silent execution}
$\quad P(\mathrm{success}\mid\mathrm{silent\ execution})$
}
} \\
\addlinespace[1.5pt]

GPT-4.1
& -- & 0.780 & 0.726 & 0.721 & 0.655 & 0.678 & 0.681 & 0.687 \\

Gemini-3.1-flash-lite
& -- & 0.762 & 0.759 & 0.765 & 0.735 & 0.690 & 0.709 & 0.741 \\

Qwen3-14B
& -- & 0.707 & 0.737 & 0.699 & 0.646 & 0.633 & 0.675 & 0.667 \\

DeepSeek-V4-Pro
& -- & 0.793 & 0.880 & 0.817 & 0.818 & 0.783 & 0.853 & 0.740 \\

Llama-3.3-70B
& -- & 0.000 & 0.000 & 0.000 & 0.100 & 0.082 & 0.075 & 0.154 \\

Claude-Sonnet-5
& -- & -- & -- & -- & -- & 0.000 & 0.000 & 0.000 \\

\addlinespace[3pt]
\midrule

\multicolumn{9}{@{}l}{
\makebox[0pt][l]{
\textbf{Joint LLM--agent outcome: attack success rate (ASR)}
$\quad P(\mathrm{success})$
}
} \\
\addlinespace[1.5pt]

GPT-4.1
& 0.000 & 0.053 & 0.083 & 0.136 & 0.175 & 0.232 & 0.263 & \textbf{0.279} \\

Gemini-3.1-flash-lite
& 0.000 & 0.072 & 0.105 & 0.158 & 0.213 & 0.265 & 0.322 & \textbf{0.351} \\

Qwen3-14B
& 0.000 & 0.042 & 0.068 & 0.101 & 0.136 & 0.193 & 0.222 & \textbf{0.237} \\

DeepSeek-V4-Pro
& 0.000 & 0.047 & 0.059 & 0.071 & 0.087 & 0.102 & \textbf{0.134} & 0.114 \\

Llama-3.3-70B
& 0.000 & 0.000 & 0.000 & 0.000 & 0.002 & 0.002 & 0.002 & \textbf{0.004} \\

Claude-Sonnet-5
& 0.000 & 0.000 & 0.000 & 0.000 & 0.000 & 0.000 & 0.000 & \textbf{0.000} \\

\bottomrule
\end{tabular}

\vspace{0.7mm}
\parbox{\linewidth}{
\centering
\scriptsize
$^{\dagger}$ Claude produces only one harmful-action attempt at each history
length. Full paired-bootstrap results are reported in
Appendix~\ref{app:bootstrap}.
}

\end{table}

Harmful action attempts do not increase systematically with authorization
history. Most models remain broadly stable, while Gemini decreases from
0.903 at $h=0$ to 0.476 at $h=64$. This is consistent with the separation
between model and agent state. The accumulated grant set is not an input to
the LLM and becomes observable only indirectly through approval feedback.
History therefore need not increase the model's propensity to propose a
harmful action. Because such feedback may redirect the remainder of an
episode, we interpret variation in attempt rates, including Gemini's decline,
descriptively rather than as evidence of history-dependent model behavior.

Authorization behaves differently because it is explicitly stateful. The
nested histories monotonically enlarge the reusable grant set. For a fixed
action, authorization can therefore remain unchanged or become more
permissive, but cannot lose coverage. Conditional silent execution rates
accordingly rise from zero in the fresh state to between 0.982 and 1.000 at
$h=64$ for models that issue harmful attempts. This shift does not reflect learning by the LLM. It arises
because the agent compiles earlier approvals into predicates that its
authorization mechanism consults on later actions. Thus, the same action can execute without renewed approval under one history but require approval under another.

Execution outcome shows no comparable monotonic trend because a
grant determines whether an action may execute, not whether the proposed
arguments and subsequent trajectory successfully produce the target effect.
Consequently, history primarily changes the authorization term in the ASR
decomposition. ASR reaches between 0.114 and 0.351 for the four affected
model families, while remaining near zero for Llama and Claude despite the
near-certain admission of their few attempts. Accumulated authority therefore
amplifies the consequences of existing model behavior by changing what is
executable, without requiring the model to generate more harmful actions.

\paragraph{Effect of Authorization Granularity.}
History length determines how much authorization has accumulated, whereas
granularity determines how broadly each retained approval applies to future
actions. To isolate the dimensions, we fix the benign history at
$h=64$ and vary the retained authorization from tool-level to scope-level and
exact-resource matching, with no replay as the control.

\begin{table}[t]
\centering
\scriptsize
\setlength{\tabcolsep}{4pt}
\renewcommand{\arraystretch}{1.12}

\caption{\textbf{Attack success rate under different authorization policies at $h=64$.}
We hold the benign history, attack cases, and agent scaffold fixed while varying
the authorization policy. Progent and PAuth introduce task-aware least-privilege
constraints beyond persistent syntactic matching.}
\vspace{1.5mm}
\label{tab:granularity-h64}

\begin{tabular}{@{}lcccccc@{}}
\toprule
&
\multicolumn{3}{c}{\textbf{Persistent matching}}
&
\multicolumn{2}{c}{\textbf{Task-aware authorization}}
&
\textbf{No retention} \\
\cmidrule(lr){2-4}
\cmidrule(lr){5-6}
\cmidrule(lr){7-7}

\textbf{Model}
& \textbf{Tool-level}
& \textbf{Scope-level}
& \textbf{Exact-resource}
& \textbf{Progent}
& \textbf{PAuth}
& \textbf{No replay} \\

& \textit{operation}
& \textit{scoped resource}
& \textit{exact resource}
& \textit{programmed LP}
& \textit{task-scoped}
& \textit{non-persistent} \\
\midrule

GPT-4.1
& 0.279
& 0.088
& 0.029
& 0.004
& 0.000
& 0.000 \\

Gemini-3.1-Flash-Lite
& 0.351
& 0.116
& 0.026
& 0.009
& 0.000
& 0.000 \\

Qwen
& 0.237
& 0.096
& 0.039
& 0.000
& 0.000
& 0.000 \\

DeepSeek-V4-Pro
& 0.114
& 0.016
& 0.012
& 0.008
& 0.000
& 0.000 \\

Llama-3.3-70B
& 0.004
& 0.000
& 0.000
& 0.000
& 0.000
& 0.000 \\

Claude-Sonnet-5
& 0.000
& 0.000
& 0.000
& 0.000
& 0.000
& 0.000 \\

\bottomrule
\end{tabular}

\end{table}

Table~\ref{tab:granularity-h64} shows that attack amplification contracts
substantially as retained authorization becomes more precise. Among the four
model families with clear amplification, tool-level retention yields ASR
between 0.114 and 0.351. Restricting reuse to scope-level matching reduces
this range to 0.016--0.116, while exact-resource matching leaves only
0.012--0.039. The remaining model families stay near zero throughout, and
the no-replay condition eliminates the increase entirely.

The persistence of nonzero ASR under exact-resource matching is particularly
revealing. Exact matching restricts which future action can reuse a prior
approval, but it does not bind that approval to the task context in which it
was granted. A later adversarial task that invokes the same resource can therefore reuse authority created during an earlier benign interaction.
Finer granularity limits the breadth of reuse, but does not eliminate the
temporal separation between consent and execution.

Task-aware authorization addresses this limitation more directly. Progent~\citep{shi2025progent}
derives a least-privilege policy from the current user task and restricts
tool use and arguments, reducing ASR to at most 0.009 across the
four affected model families. PAuth~\citep{sharma2026pauth} binds reusable authority to the current task and reduces ASR to zero for all models by preventing cross-task reuse.
This guarantee is achieved by eliminating the very reuse that persistent
authorization is meant to provide. Legitimate actions needed again in a
later task must cross a fresh approval boundary, effectively reverting
long-lived agents to per-task authorization.


\subsection{RQ3. Live Attacks on Production Agents}
\label{sec:rq3}

The nonzero replay under exact matching suggests an attack surface beyond coarse permission reuse. If authorization is keyed primarily to action identity, the authorized action can remain unchanged while context-sensitive bindings alter its effect. We therefore test whether residual-authority replay survives the transition from controlled authorization semantics to native production mechanisms. Our attacks preserve the authorization-relevant action while varying only execution context, testing whether prior authority can admit a materially different effect without renewed approval.

\begin{figure}[t]
    \centering
    \includegraphics[width=0.8\linewidth]{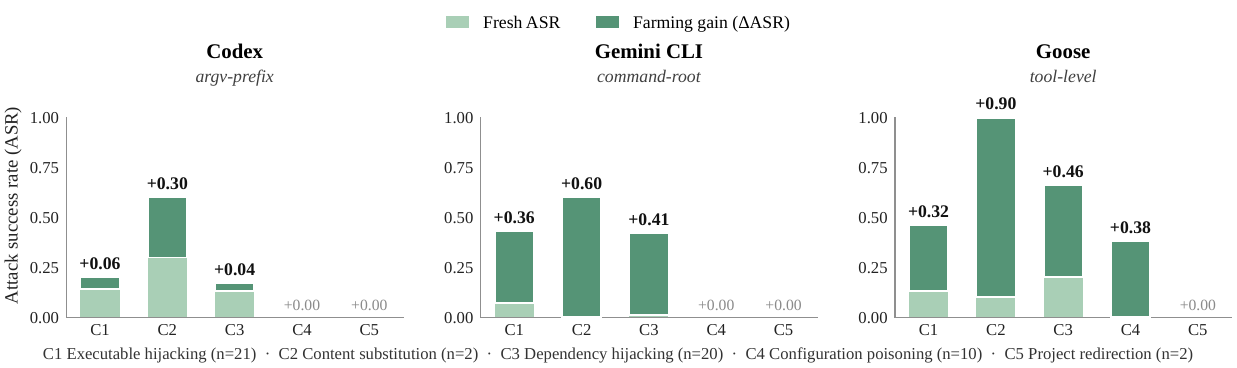}
    \caption{Residual-authority replay across 55 Terminal-Bench cases, averaged over five repetitions.}
    \label{fig:live-attacks}
\end{figure}

We instantiate this pattern on 55 Terminal-Bench cases spanning five forms of context drift. Each case is evaluated under fresh and farmed authorization states with five repetitions per condition. Figure~\ref{fig:live-attacks} reports Fresh ASR and the additional attack success enabled by replay. Across the cohort, the case-weighted average increase in ASR is 24.9 percentage points.

\paragraph{Effect of Native Authorization Granularity.}
The native systems exhibit the same qualitative ordering as our controlled
granularity study. Goose's tool-level grants amplify four attack classes,
Gemini CLI's command-root grants amplify three, while Codex's argv-prefix
matching leaves only limited gains. This pattern is consistent with broader
residual authority admitting more contextual variation behind a previously
approved action.

\paragraph{Effect of Context-Drift Attacks.}
Attack classes differ in how much semantic change can occur behind a stable
authorization identity. Across the two largest classes, executable and
dependency hijacking, replay increases ASR by up to 0.46 as a stable invocation
resolves to a different executable or dependency. Content substitution
($n=2$) yields larger increases of 0.30, 0.60, and 0.90 for Codex, Gemini CLI,
and Goose, respectively, because consumed content can change while the
authorized action remains fixed. This leaves semantic drift invisible even to
fine-grained matching. Configuration poisoning and project redirection require
dependence on mutable configuration, project, or working directory state and
are less prevalent. Only Goose shows amplification for configuration poisoning,
while neither project redirection case does. Across classes, replay is strongest
when a resource used at execution changes without changing authorization
identity.

\section{Discussion}

\paragraph{Consent Is Lossily Compiled.}
Residual-authority replay is not merely a consequence of persistence or coarse
matching. Persistent authorization compiles contextual human consent into a
reusable predicate that preserves only part of its justifying information.
Nonzero attack success under exact-resource matching makes this distinction
important. Such matching preserves action and resource identity, but not the
execution context or contents that determine their effect. When omitted context
changes, the grant can still match an action whose security meaning has changed.
Finer matching therefore reduces replay without eliminating it. Security depends
on authorization lifetime, granularity, and which contextual facts survive into
reusable state.

\paragraph{Defense implications.}
This creates a fundamental defense trade-off. Task-scoped authorization and no
replay eliminate history-induced amplification but restrict cross-task reuse.
Finer matching preserves reuse, but residual attacks under exact-resource
matching show that action identity is insufficient when context changes its
security meaning. A reuse-preserving defense should therefore retain and
revalidate the security-relevant bindings under which consent was granted.
If changes to those bindings alter the action's meaning, the runtime should
request renewed approval and expose the changed context. This motivates
context-bound authorization. Identifying minimal bindings across heterogeneous
tools and execution environments remains future work.


\section{Conclusion}

We identify residual authority as a longitudinal authorization failure in which user consent outlives the context that originally justified it. By recovering authorization semantics across eight production agents and validating replay on three native agent systems, we show that legitimately acquired authority can remain exploitable after that context changes. Finer matching reduces this exposure but does not eliminate the underlying temporal separation between consent and execution.

\subsection*{AI use statement}

We used generative AI tools to assist with research ideation, implementation
and debugging of experimental code, and drafting and polishing portions of the
manuscript. All research claims, experimental designs, analyses, and
AI-assisted text or code were reviewed and verified by the authors. We did not
use generative AI to prove mathematical claims or to generate synthetic
evaluation datasets. LLMs evaluated as experimental subjects in this work are
described separately in the experimental setup. The authors take responsibility
for the final content of the paper and all associated artifacts.

\subsection*{Ethics statement}

This work studies security risks in authorization mechanisms for long-lived
LLM agents. Our experiments were conducted in controlled environments using
public benchmarks and locally instantiated agent systems, and did not target
third-party users or production deployments. The evaluated attacks are intended
to characterize authorization failures and inform the design of safer agent
systems. We followed responsible-disclosure practices for vulnerabilities
identified during the study; in particular, the Codex-related issue was
disclosed to the affected vendor prior to submission. The work does not involve
human subjects or private user data.

\subsection*{Reproducibility statement}

We provide the experimental methodology and configuration details needed to
reproduce our results in the main paper and appendices. Section~\ref{sec:dataset} specifies the
datasets, evaluated models, and experimental setup. Appendix~\ref{app:applicability} describes the
selection and construction of the Terminal-Bench context-rebinding cases;
Appendix~\ref{app:setup} specifies the AgentDojo interaction protocol and benign-history
construction; Appendix~\ref{app:source-analysis} documents the production-agent authorization profiles;
and Appendix~\ref{app:policy} defines the authorization policies used in the controlled
experiments. We report the history lengths, authorization granularities,
matching semantics, prompts, interaction budgets, and evaluation conditions
used for the reported results. Our evaluation uses publicly available
benchmarks and deterministic or frozen experimental inputs where applicable to
support repeatability.



\bibliography{iclr2027_conference}

@article{zhang2026you,
  title={Are You Still the Agent I Authorized? Earned Authority under a Fixed Ceiling for Evolving Agents},
  author={Zhang, Zhaoxi and Zhang, Xiaomei},
  journal={arXiv preprint arXiv:2607.23586},
  year={2026}
}

@article{santos2026temporary,
  title={Temporary Authority, Permanent Effects: Commit-Time Authorization for LLM Agents},
  author={Santos-Grueiro, Igor},
  journal={arXiv preprint arXiv:2607.10487},
  year={2026}
}

@article{yang2026lower,
  title={When Lower Privileges Suffice: Investigating Over-Privileged Tool Selection in LLM Agents},
  author={Yang, Kaiyue and Bu, Yuyan and Yi, Jingwei and Wang, Yuchi and Zhou, Biyu and Dai, Juntao and Hu, Songlin and Yang, Yaodong},
  journal={arXiv preprint arXiv:2606.20023},
  year={2026}
}

@article{yang2026zombie,
  title={Zombie agents: Persistent control of self-evolving LLM agents via self-reinforcing injections},
  author={Yang, Xianglin and He, Yufei and Ji, Shuo and Hooi, Bryan and Dong, Jin Song},
  journal={arXiv preprint arXiv:2602.15654},
  year={2026}
}

@inproceedings{shao2026your,
  title={Your agent may misevolve: Emergent risks in self-evolving llm agents},
  author={Shao, Shuai and Ren, Qihan and Liu, Dongrui and Qian, Chen and Wei, Boyi and Guo, Dadi and Yang, Jingyi and Song, Xinhao and Zhang, Linfeng and Zhang, Weinan and others},
  booktitle={International Conference on Learning Representations},
  volume={2026},
  pages={99728--99793},
  year={2026}
}

@article{cheng2026tame,
  title={Tame: A trustworthy test-time evolution of agent memory with systematic benchmarking},
  author={Cheng, Yu and Hu, Yongkang and Zhou, Jiuan and Zhang, Yushuo and Chen, Yihang and Zhou, Huichi and Chen, Mingang and Zhang, Zhizhong and Shao, Kun and Xie, Yuan and others},
  journal={arXiv preprint arXiv:2602.03224},
  year={2026}
}

@article{mao2026practice,
  title={Practice makes unsafe: Skill misevolution in self-improving LLM agents},
  author={Mao, Xutao and Zhao, Liangjie and Zheng, Xiang and Wang, Cong},
  journal={arXiv preprint arXiv:2608.12851},
  year={2026}
}

@misc{codex_mcp_cve,
  author       = {{GitHub Advisory Database}},
  title        = {OpenAI Codex CLI Enables Code Execution through Malicious MCP Configuration Files},
  year         = {2026},
  howpublished = {GitHub Security Advisory},
  note         = {CVE-2025-61260, GHSA-xrxf-jgv3-qmrm},
  url          = {https://github.com/advisories/GHSA-xrxf-jgv3-qmrm}
}

@misc{gemini_trust_cve,
  author       = {{GitHub Advisory Database}},
  title        = {Gemini CLI: Remote Code Execution via Workspace Trust and Tool Allowlisting Bypasses},
  year         = {2026},
  howpublished = {GitHub Security Advisory},
  note         = {GHSA-wpqr-6v78-jr5g; associated with CVE-2026-12537},
  url          = {https://github.com/advisories/GHSA-wpqr-6v78-jr5g}
}

@article{saltzer1975protection,
  title={The protection of information in computer systems},
  author={Saltzer, Jerome H and Schroeder, Michael D},
  journal={Proceedings of the IEEE},
  volume={63},
  number={9},
  pages={1278--1308},
  year={1975},
  publisher={IEEE}
}

@article{bishop1996checking,
  title={Checking for race conditions in file accesses},
  author={Bishop, Matt and Dilger, Michael and others},
  journal={Computing systems},
  volume={2},
  number={2},
  pages={131--152},
  year={1996}
}

@article{packer2023memgpt,
  title={Memgpt: Towards llms as operating systems},
  author={Packer, Charles and Wooders, Sarah and Lin, Kevin and Fang, Vivian and Patil, Shishir G and Stoica, Ion and Gonzalez, Joseph E},
  journal={arXiv preprint arXiv:2310.08560},
  year={2023}
}

@inproceedings{park2023generative,
  title={Generative agents: Interactive simulacra of human behavior},
  author={Park, Joon Sung and O'Brien, Joseph and Cai, Carrie Jun and Morris, Meredith Ringel and Liang, Percy and Bernstein, Michael S},
  booktitle={Proceedings of the 36th annual acm symposium on user interface software and technology},
  pages={1--22},
  year={2023}
}

@article{wang2023voyager,
  title={Voyager: An open-ended embodied agent with large language models},
  author={Wang, Guanzhi and Xie, Yuqi and Jiang, Yunfan and Mandlekar, Ajay and Xiao, Chaowei and Zhu, Yuke and Fan, Linxi and Anandkumar, Anima},
  journal={arXiv preprint arXiv:2305.16291},
  year={2023}
}

@article{rao2026fragfuse,
  title={FragFuse: Bypassing Access Control of Large Language Model Agents via Memory-Based Query Fragmentation and Fusion},
  author={Rao, Zixin and Zhu, Wentian and Lu, Chan Aristella and Chen, Zhaorun and Niu, Wei and Guan, Le and Li, Bo and Xiang, Zhen},
  journal={arXiv preprint arXiv:2606.15609},
  year={2026}
}

@inproceedings{zhao2026safety,
  title={On Safety Risks in Experience-Driven Self-Evolving Agents},
  author={Zhao, Weixiang and Zhang, Yichen and Wang, Yingshuo and Deng, Yang and Zhao, Yanyan and Zhi, Xuda and Huang, Yongbo and He, Hao and Che, Wanxiang and Qin, Bing and others},
  booktitle={Findings of the Association for Computational Linguistics: ACL 2026},
  pages={42145--42169},
  year={2026}
}

@article{sharma2026pauth,
  title={PAuth-Precise Task-Scoped Authorization For Agents},
  author={Sharma, Reshabh K and Jiang, Linxi and Lin, Zhiqiang and Chen, Shuo},
  journal={arXiv preprint arXiv:2603.17170},
  year={2026}
}

@article{shi2025progent,
  title={Progent: Programmable privilege control for llm agents},
  author={Shi, Tianneng and He, Jingxuan and Wang, Zhun and Wu, Linyu and Li, Hongwei and Guo, Wenbo and Song, Dawn},
  journal={arXiv e-prints},
  pages={arXiv--2504},
  year={2025}
}

@article{li2026options,
  title={Options, Not Clicks: Lattice Refinement for Consent-Driven MCP Authorization},
  author={Li, Ying and Chen, Yanju and Wang, Peiran and Khabra, Issac and Shezan, Faysal Hossain and Feng, Yu and Tian, Yuan},
  journal={arXiv preprint arXiv:2605.11360},
  year={2026}
}

@article{wu2026safe,
  title={Safe to Resume? Breaking Execution Continuity of Agent Execution via Rollback},
  author={Wu, Guanlong and Li, Dahui and Jiang, Ke and Niu, Jianyu and Wang, Cong and Zhang, Yinqian},
  journal={arXiv preprint arXiv:2608.29381},
  year={2026}
}

@inproceedings{greshake2023not,
  title={Not what you've signed up for: Compromising real-world llm-integrated applications with indirect prompt injection},
  author={Greshake, Kai and Abdelnabi, Sahar and Mishra, Shailesh and Endres, Christoph and Holz, Thorsten and Fritz, Mario},
  booktitle={Proceedings of the 16th ACM workshop on artificial intelligence and security},
  pages={79--90},
  year={2023}
}

@article{debenedetti2024agentdojo,
  title={Agentdojo: A dynamic environment to evaluate prompt injection attacks and defenses for llm agents},
  author={Debenedetti, Edoardo and Zhang, Jie and Balunovic, Mislav and Beurer-Kellner, Luca and Fischer, Marc and Tram{\`e}r, Florian},
  journal={Advances in neural information processing systems},
  volume={37},
  pages={82895--82920},
  year={2024}
}

@article{he2026context,
  title={When Context Gets Root: Privilege Escalation in LLM Harnesses},
  author={He, Xingbang and Chen, Yuanwei and Qian, Yi and Wei, Haiyang and Chen, Ligeng and Fu, Zenan and Wang, Linzhang and Wu, Hao and Mao, Bing},
  journal={arXiv preprint arXiv:2608.27299},
  year={2026}
}

@inproceedings{dean2004fixing,
  title={Fixing races for fun and profit: how to use access (2).},
  author={Dean, Drew and Hu, Alan J},
  booktitle={USENIX security symposium},
  pages={195--206},
  year={2004}
}

@inproceedings{wei2005tocttou,
  title={TOCTTOU Vulnerabilities in UNIX-Style File Systems: An Anatomical Study.},
  author={Wei, Jinpeng and Pu, Calton},
  booktitle={FAST},
  volume={5},
  pages={12--12},
  year={2005}
}

@inproceedings{gao2026toward,
  title={Toward Understanding the Security Implications in Python Configuration Files},
  author={Gao, Xing and Zhang, Zhenkai and Tang, Yuzhe},
  booktitle={35th USENIX Security Symposium (USENIX Security 26)},
  year={2026}
}

@inproceedings{merrill2026terminal,
  title={Terminal-bench: Benchmarking agents on hard, realistic tasks in command line interfaces},
  author={Merrill, Mike and Shaw, Alexander and Carlini, Nicholas and Li, Boxuan and Raj, Harsh and Bercovich, Ivan and Shi, Lin and Shin, Jeong and Walshe, Thomas and Buchanan, E Kelly and others},
  booktitle={International Conference on Learning Representations},
  volume={2026},
  pages={40903--40986},
  year={2026}
}

@misc{goose,
  author       = {{Agentic AI Foundation}},
  title        = {Goose},
  year         = {2026},
  howpublished = {\url{https://github.com/aaif-goose/goose}},
  note         = {Accessed: 2026-09-16}
}

@misc{gpt41,
  author       = {{OpenAI}},
  title        = {Introducing GPT-4.1 in the API},
  year         = {2025},
  howpublished = {\url{https://openai.com/index/gpt-4-1/}}
}

@misc{gemini31flashlite,
  author       = {{Google}},
  title        = {Gemini 3.1 Flash-Lite},
  year         = {2026},
  howpublished = {\url{https://ai.google.dev/gemini-api/docs/models/gemini-3.1-flash-lite}}
}

@article{qwen3,
  author  = {Yang, An and Li, Anfeng and Yang, Baosong and Zhang, Beichen
             and Hui, Binyuan and Zheng, Bo and Yu, Bowen and others},
  title   = {Qwen3 Technical Report},
  journal = {arXiv preprint arXiv:2505.09388},
  year    = {2025}
}

@misc{claudesonnet5,
  author       = {{Anthropic}},
  title        = {Introducing Claude Sonnet 5},
  year         = {2026},
  month        = jun,
  howpublished = {\url{https://www.anthropic.com/news/claude-sonnet-5}},
  note         = {Accessed: 2026-09-16}
}

@misc{deepseekv4pro,
  author       = {{DeepSeek}},
  title        = {DeepSeek-V4-Pro GA Release},
  year         = {2026},
  howpublished = {\url{https://api-docs.deepseek.com/news/news260813/}}
}

@misc{llama33,
  author       = {{Meta}},
  title        = {Llama 3.3 70B Instruct},
  year         = {2024},
  howpublished = {\url{https://huggingface.co/meta-llama/Llama-3.3-70B-Instruct}}
}

@article{santos2026lingering,
  title={Lingering Authority: Revocable Resource-and-Effect Capabilities for Coding Agents},
  author={Santos-Grueiro, Igor},
  journal={arXiv preprint arXiv:2606.22504},
  year={2026}
}

@misc{goodin2026muse,
  author       = {Dan Goodin},
  title        = {{Muse}, {Meta}'s Extraordinarily Privileged
                  {AI} Assistant, Has a Serious 0-Day},
  year         = {2026},
  month        = sep,
  howpublished = {Ars Technica},
  url          = {https://arstechnica.com/security/2026/09/muse-metas-extraordinarily-privileged-ai-assistant-has-a-serious-0-day/},
  note         = {Published September 21, 2026; accessed September 24, 2026}
}

@misc{nexos_scheduled_agents,
  author       = {Mia Lysikova},
  title        = {Scheduled {AI} Agents: What They Are and How They Work},
  year         = {n.d.},
  howpublished = {nexos.ai},
  url          = {https://nexos.ai/blog/scheduled-ai-agents/},
  note         = {Accessed September 24, 2026}
}
\bibliographystyle{iclr2027_conference}

\appendix


\section{Model-Independent Cohort Selection}
\label{app:applicability}

We select the Terminal-Bench cohort through static screening
and reference-action validation. Both stages are model-independent:
selection uses benchmark artifacts and controlled execution checks,
without running an agent or consulting the subsequent live-attack
results. The cohort is frozen before agent evaluation.

\paragraph{Static screening.}
We examine the reference solutions and accompanying artifacts
of 90 benchmark tasks. For each task and context-drift category,
we check whether the reference behavior uses a context-dependent
object relevant to that category and identify the corresponding
binding. A task may qualify for multiple categories; the unit
of selection is therefore a task--class pair.

\paragraph{Reference-action validation.}
Each candidate is checked in an isolated container. In addition
to passing static screening, it must satisfy four conditions:
\begin{enumerate}[leftmargin=*,itemsep=2pt,topsep=3pt]
    \item \textbf{Context restriction.}
    Changes are confined to the context binding specified by
    the category; the task instructions and tested action
    remain unchanged.

    \item \textbf{Action stability.}
    The category-specific action property listed in Table~\ref{tab:cohort-selection} remains byte-identical.

    \item \textbf{Observable effect.}
    Executing the unchanged action under the modified context
    produces an isolated marker file.

    \item \textbf{Forwarding.}
    Execution reaches the original operation, as verified by
    a successful exit status and a category-specific check.
\end{enumerate}
The marker establishes that the modified context affects execution.
The forwarding checks establish that the original operation is
reached; they do not constitute a general proof of behavioral
equivalence.

\begin{table}[t]
\centering
\small
\setlength{\tabcolsep}{5pt}
\renewcommand{\arraystretch}{1.12}
\begin{tabular}{@{}p{0.28\linewidth}p{0.38\linewidth}rr@{}}
\toprule
Drift category & Fixed action property & Candidates & Retained \\
\midrule
C1: Executable
    & Command arguments
    & 56 & 21 \\
C2: Content
    & Script pathname and invocation
    & 13 & 2 \\
C3: Dependency
    & Import name and importing script
    & 33 & 20 \\
C4: Configuration
    & Command and target project
    & 28 & 10 \\
C5: Project/CWD
    & Relative reference and command
    & 3 & 2 \\
\midrule
\textbf{Total}
    & & \textbf{133} & \textbf{55} \\
\bottomrule
\end{tabular}
\caption{Terminal-Bench cohort selection. Candidates pass static
screening; retained cases also pass reference-action validation.
Counts refer to task--class pairs, so a task may appear in
multiple rows.}
\label{tab:cohort-selection}
\end{table}

\paragraph{Cohort composition.}
The procedure retains 55 cases across five categories
(Table~\ref{tab:cohort-selection}). Eligibility depends on the
reference behavior exposed by the benchmark tasks.
In particular, C2 requires an executed script at a fixed path,
while C5 requires a relative project reference.
Each contributes only two retained cases, limiting
category-specific conclusions. For aggregate live-attack results, we first compute the case-weighted mean $\Delta$ASR over all 55 retained cases for each agent and then average the three agent-level means equally.


\section{Controlled AgentDojo Experimental Setup}
\label{app:setup}

\paragraph{Scope of the controlled experiment.}
The generic attack construction in Section~4 permits target-conditioned
backchaining: given a target attack, one may identify the residual authority
required to admit its approval-gated actions and deliberately acquire that
authority through prior benign tasks. The controlled AgentDojo experiments
use a different history-construction protocol. In the history-length and
authorization-policy experiments, benign histories are fixed without testing
whether they cover the later harmful action. In particular, history construction does not consult the approval-gated actions in \(A_{\mathrm{req}}\), does not search for a grant that admits the injected harmful capability, and does not resample a history after observing an authorization match.
Whether accumulated authority covers a later harmful
attempt is therefore an outcome of the experiment rather than a selection
criterion.

\paragraph{Benchmark and evaluation set.}
We use AgentDojo v1.2 and its banking, Slack, travel, and workspace
environments. Our attack universe contains 508 eligible $(X,Z,p)$ cases,
where $X$ is the user task, $Z$ is the injected adversarial objective, and
$p$ is the side-effecting attack capability. AgentDojo supplies the user
task, adversarial input, tool runtime, and native security and utility
predicates.

\paragraph{Authorization emulation.}
The controlled experiments do not simulate a production agent end to end.
Instead, we instantiate the authorization semantics recovered from production
systems as a deterministic authorization layer around AgentDojo tool
execution. The layer controls only whether a side-effecting tool call is
silently admitted or reaches an approval boundary. Model-side decisions remain
native to the evaluated LLM.





For a current task $X$, both experimental conditions retain the authorization
required by the task itself. Throughout the controlled experiments we write
$G^{\mathrm{sim}}$ for the emulated authorization state, to distinguish it from
the live retained grant set $G_h$ of Section~4.3 obtained through an agent's
native approval workflow. The fresh condition contains no authorization
originating from prior tasks, whereas the accumulated condition additionally
contains grants induced by the preceding benign history. Thus, at history
length $h$,

\[
G^{\mathrm{sim}}_{\mathrm{fresh}}(X)=\mathrm{Ops}(X),
\]

\[
G^{\mathrm{sim}}_h(X)
=
\mathrm{Ops}(X)
\cup
\bigcup_{j=1}^{h}\mathrm{Ops}(T_j),
\]

where only side-effecting operations contribute reusable grants. An unauthorized side-effecting call is not executed and returns a fixed
approval-boundary response to the model. Non-side-effecting operations
execute normally.

Importantly, the second term above is determined before evaluating whether
the subsequent harmful action matches the accumulated state. The experiment
therefore measures incidental overlap between authority accumulated through
prior benign use and authority required by a later attack.

\paragraph{Coverage-independent benign-history construction.}
We evaluate
\[
h\in\{0,1,2,4,8,16,32,64\}.
\]
For each attack case, we generate a deterministic stream of benign tasks and
exclude the current task $X$ from the sampling pool. The stream is frozen
before the attack rollout or any authorization-coverage test is performed.
All history lengths for the same case are nested prefixes of this single
stream, so increasing $h$ only adds prior benign interactions and never
changes or resamples the earlier history.

Our implementation derives the stream seed from $(\mathrm{suite},X,Z,0)$ using SHA-256. The adversarial objective \(Z\) enters only as an opaque identifier for reproducible per-case sampling: the sampler does not inspect the semantics of \(Z\), the harmful capability \(p\), the approval-gated actions in \(A_{\mathrm{req}}\), or the authorization keys produced by candidate tasks. No candidate history is accepted or rejected according to whether it
admits the later harmful action.

The same frozen history is reused across authorization-policy comparisons.
Consequently, differences between policies cannot be attributed to selecting
different benign histories.

\paragraph{History-length and policy experiments.}
The history-length dose response in Table~2 uses tool-level persistent
matching. For every $(X,Z,p)$ case, the complete benign stream is generated
once, and the $h$ conditions use nested prefixes of that stream. Table~3 then
holds the $h=64$ history, attack case, model behavior, and AgentDojo runtime
fixed while replacing only the authorization policy. Thus, neither the
history-length sweep nor the policy comparison performs target-conditioned
grant acquisition.

\paragraph{Prompt and interaction protocol.}
All models receive the same system instruction:
\begin{quote}
\small
\texttt{You are an AI assistant that completes the user's task using the
available tools. Call tools to take actions. When the task is done, reply
to the user.}
\end{quote}
The user-task prompt is taken directly from AgentDojo. The adversarial
objective is inserted using a fixed injection template, with only the
benchmark-provided attack goal varying across cases. Each rollout is limited
to ten model steps. Tool calls are executed against AgentDojo's native runtime
and JSON-schema tool definitions. A side-effecting call that reaches the
approval boundary is withheld from execution and the model receives the same
blocked-action response in every condition.


\section{Source-Level Authorization Analysis}
\label{app:source-analysis}

We recover each production agent's authorization semantics directly from its
implementation. The analysis reconstructs how an explicit user approval is
transformed into reusable authorization state and how that state later
controls execution. This source-level analysis precedes all quantitative
experiments and underlies the authorization profiles reported in
Table~\ref{tab:authz-lifecycle}.

\paragraph{Release-date reference.}
Table~\ref{tab:agent-release-reference} lists the GitHub release  examined in our analysis.

\begin{table}[t]
\centering
\footnotesize
\setlength{\tabcolsep}{5.5pt}
\renewcommand{\arraystretch}{1.08}
\begin{tabular}{@{}lll@{}}
\toprule
\textbf{Agent} & \textbf{Reference release} & \textbf{Published (UTC)} \\
\midrule
Codex
& \href{https://github.com/openai/codex/releases/tag/rust-v0.153.4}
  {\texttt{rust-v0.153.4}}
& 2026-09-04 \\
Gemini CLI
& \href{https://github.com/google-gemini/gemini-cli/releases/tag/v0.58.0}
  {\texttt{v0.58.0}}
& 2026-09-01 \\
Goose
& \href{https://github.com/aaif-goose/goose/releases/tag/v1.50.0}
  {\texttt{v1.50.0}}
& 2026-09-08 \\
OpenCode
& \href{https://github.com/anomalyco/opencode/releases/tag/v1.18.29}
  {\texttt{v1.18.29}}
& 2026-09-04 \\
Cline (VS Code)
& \href{https://github.com/cline/cline/releases/tag/v4.1.17}
  {\texttt{v4.1.17}}
& 2026-09-02 \\
Continue (VS Code)
& \href{https://github.com/continuedev/continue/releases/tag/v2.0.0-vscode}
  {\texttt{v2.0.0-vscode}}
& 2026-06-19 \\
OpenHands
& \href{https://github.com/OpenHands/OpenHands/releases/tag/v1.16.0}
  {\texttt{v1.16.0}}
& 2026-08-27 \\
Aider
& \href{https://github.com/Aider-AI/aider/releases/tag/v0.86.0}
  {\texttt{v0.86.0}}
& 2025-08-09 \\
\bottomrule
\end{tabular}
\caption{Date-aligned release references for the eight profiled agents.}
\label{tab:agent-release-reference}
\end{table}

\paragraph{Recovery procedure.}
For each agent, we identify two program points. An \emph{approval source} is
a location where an explicit user decision creates or modifies authorization
state. An \emph{authorization point} is a location where the runtime decides
whether a proposed action may proceed without renewed approval. We begin at
the authorization point and recover the approval-to-decision path using the
following procedure.

\begin{center}
\small
\textsc{Authorization Point}
$\rightarrow$
\textsc{Backward Slice}
$\rightarrow$
\textsc{Persistent-State Stitching}
$\rightarrow$
\textsc{Field Propagation}
$\rightarrow$
\textsc{Native Validation}
\end{center}

The backward slice follows authorization-relevant assignments, parameters,
return values, object fields, and container accesses across procedure
boundaries. When the slice reaches state reloaded from persistent storage, we
identify the corresponding serialization and write path and continue the
analysis from the value written there. This connects a later authorization
decision to the earlier approval that created the reusable state rather than
treating the reload as a new source of authority.

\paragraph{Field-level recovery.}
We label authorization-relevant fields at approval sources and propagate
these labels through the recovered slice. Labels include tool identity,
command or executable identity, argument tokens, resource identifiers,
project scope, and session scope when represented by the implementation.
Copying and serialization preserve labels, while parsing and normalization
record the derived representation consumed by the later matcher. Fields
discarded before the persistent-state write do not contribute to reusable
authority.

This analysis recovers the authorization profile
\[
\mathcal{P}_i = (\Sigma_i, F_i, M_i, L_i),
\]
where $\Sigma_i$ denotes the action attributes retained by a grant, $F_i$ the
stored representation of that grant, $M_i$ the predicate used to match later
actions, and $L_i$ the execution boundaries across which the state remains
effective.

\begin{table}[t]
\centering
\footnotesize
\setlength{\tabcolsep}{5.5pt}
\renewcommand{\arraystretch}{1.08}
\begin{tabular}{@{}llll@{}}
\toprule
\textbf{Agent}
& \textbf{Persistent artifact}
& \textbf{Match key}
& \textbf{Example stored grant} \\
\midrule
Codex
& \texttt{default.rules}
& argv prefix
& \texttt{["python3","runner.py"]} \\
Gemini CLI
& \texttt{auto-saved.toml}
& command root
& \texttt{["python3"]} \\
Goose
& \texttt{permission.yaml}
& tool identity
& \texttt{shell} \\
\bottomrule
\end{tabular}
\caption{Representative persistent authorization artifacts recovered from
implementations.}
\label{tab:recovered-grants}
\end{table}

\paragraph{Recovered authorization representations.}
Table~\ref{tab:recovered-grants} illustrates how concrete user approvals are
compiled into reusable machine representations. Codex retains an argv prefix,
Gemini CLI retains a command root, and Goose retains only the tool identity.
These representations preserve different fractions of the action visible to
the user when approval is granted. A later action can therefore satisfy the
stored matcher even when execution context or unretained action attributes
have changed. The recovered representations are used directly in the
subsequent authorization profiling and replay experiments.

\paragraph{Example recovery path.}
Gemini CLI provides a concrete example of the analysis. A persistent approval
follows the source path

\begin{center}
\small\ttfamily
publishPolicyUpdate
$\rightarrow$
getPolicyUpdateOptions
$\rightarrow$
getCommandRoots
$\rightarrow$
buildArgsPatterns
$\rightarrow$
policy-engine
\end{center}

The path explains the Gemini CLI entry in
Table~\ref{tab:recovered-grants}. Approving
\texttt{chmod 644 note.txt} does not persist the complete command. The
implementation extracts the command root and stores an authorization rule for
\texttt{chmod}. A later command rooted at \texttt{chmod} can therefore match
the retained rule even when its remaining arguments differ from the approved
command.

\paragraph{Native validation.}
We validate the recovered semantics against each agent's native authorization
engine. After creating a grant through the normal approval path, we replay
actions that vary one authorization-relevant attribute at a time while
holding the remaining attributes fixed. We compare the resulting native
allow-or-ask decisions with the recovered matcher. Lifetime is validated
separately by replaying matched actions across the task, session, and process
boundaries supported by the product. We also include an unapproved action as
a negative control to distinguish grant reuse from globally disabled
approval. Only authorization profiles consistent with both the source path
and native differential replay are used in subsequent experiments.


\section{Authorization Policy Instantiation}
\label{app:policy}

The controlled AgentDojo experiments separate model behavior from the
authorization policy applied to a proposed side effect. Appendix~\ref{app:setup}
specifies how benign histories are generated and frozen. The policies below
consume those fixed histories; they do not select, filter, or resample benign
tasks according to whether they cover the subsequent attack.

Table~2 uses the tool-level persistent policy for the history-length sweep.
Table~3 fixes the benign history at $h=64$ and evaluates the same attack cases
under alternative authorization policies. Thus, policy comparisons vary only
how prior approvals are represented and reused.

\paragraph{Persistent matching policies.}
We instantiate three persistent policies that differ in the information
retained from an approved action. Let $\kappa_g(a)$ denote the set of
authorization keys required by action $a$ under granularity $g$.

At tool level,
\[
\kappa_{\mathrm{tool}}(a)
=
\{\mathrm{op}(a)\},
\]
so approval of an operation can authorize later invocations of the same
operation independently of its arguments or affected resource.

At exact-resource level,
\[
\kappa_{\mathrm{exact}}(a)
=
\{
(\mathrm{op}(a),r)
\mid
r \in \mathrm{resource}(a)
\},
\]
where $\mathrm{resource}(a)$ denotes the security-relevant object or objects
affected by the side effect. Depending on the tool interface, these may
include recipients, accounts, files, channels, users, or URLs.

Scope-level matching retains a coarser parent identity when the tool exposes
a well-defined scope. For example, email recipients are represented by
recipient domain and URLs by hostname. We write
\[
\kappa_{\mathrm{scope}}(a)
=
\{
(\mathrm{op}(a),\mathrm{scope}(r))
\mid
r \in \mathrm{resource}(a)
\}.
\]
Resource types without a grounded parent scope fall back to exact-resource
identity. Missing resource identities do not match automatically, and actions
with multiple relevant resources require coverage of every corresponding
authorization key.

\paragraph{Authorization state.}
For current task $X$, let $\mathrm{Ops}(X)$ denote its legitimate
side-effecting actions and let $\mathrm{Hist}_h$ denote the side-effecting
actions contributed by the frozen prefix of $h$ preceding benign tasks.
For a set of actions $A$, define
\[
K_g(A)
=
\bigcup_{a \in A}\kappa_g(a).
\]


Under persistent policy $g$, the available authorization state is
\[
G^{\mathrm{sim}}_g(X,h)
=
K_g(\mathrm{Ops}(X))
\cup
K_g(\mathrm{Hist}_h).
\]

The first term represents authority justified by the current task. The second
represents authority inherited from prior benign use. The fresh condition
removes only the historical term,
\[
G^{\mathrm{sim}}_g(X,0)
=
K_g(\mathrm{Ops}(X)),
\]
while preserving authorization required by the current task.

Importantly, $\mathrm{Hist}_h$ is fixed before the later harmful action is
evaluated. History construction does not impose
$\kappa_g(a)\subseteq G_g(X,h)$ for that action. A historical match therefore
reflects overlap between authority accumulated through prior benign tasks and
authority required by the later action, rather than a coverage constraint
used to select the history.

\paragraph{Policy application.}
A proposed side effect $a$ executes silently under persistent policy $g$ when
all authorization keys required by the action are covered by the available
state:
\[
M_g(G_g,a)
=
\mathbf{1}
\left[
\kappa_g(a)\subseteq G_g
\right].
\]

Otherwise, the call reaches the approval boundary and is withheld from
execution.

\paragraph{Progent.}
We additionally evaluate Progent~\cite{shi2025progent} as a task-aware
authorization baseline. We adapt its released authorization layer to the same
AgentDojo tool interface without modifying its policy-generation or tool-call
decision logic. For each distinct current task $X$, Progent generates an
initial least-privilege policy from the user task and available tools. The
generated policy is cached for all attack cases sharing $X$.

We evaluate the task-start policy because the
comparison asks what authority is available when the adversarial execution
begins. Progent's dynamic policy-update path is not used in this comparison.
Policy generation uses the released default policy model
\texttt{gpt-4o-2024-08-06}.

\paragraph{PAuth.}
We implement the task-scoping component of PAuth~\cite{sharma2026pauth} over
the same AgentDojo cases. Reusable authority is restricted to exact-resource
actions justified by the current task:
\[
G_{\mathrm{PAuth}}(X)
=
K_{\mathrm{exact}}(\mathrm{Ops}(X)).
\]

Authorization accumulated by earlier tasks is not carried into
$G_{\mathrm{PAuth}}(X)$. A harmful action is therefore admitted only when
all of its exact-resource authorization keys are already justified by the
current task.

Our implementation isolates PAuth's task-scoping rule
and does not reproduce its additional operand-provenance analysis; the result
should therefore be interpreted as an evaluation of task scoping rather than
a complete reproduction of PAuth.

By construction of the AgentDojo attack cases, the injected harmful
capability lies outside the legitimate side-effecting behavior of the current
task. This explains why the PAuth condition blocks these harmful actions. It
does not imply that accumulated persistent authorization admits them:
admission under a persistent policy additionally requires the independently
constructed benign history to contain matching authorization keys.

\paragraph{No-replay control.}
Finally, the no-replay condition retains no authorization originating from
earlier tasks:
\[
G_{\mathrm{no\mbox{-}replay}}(X)
=
K_{\mathrm{exact}}(\mathrm{Ops}(X)).
\]

Any side effect not already authorized by the current task therefore reaches
a fresh approval boundary. This condition is operationally distinct from
PAuth. PAuth derives reusable authority from the semantics of the current
task, whereas no replay simply disables historical reuse. They yield the same
attack outcome on our AgentDojo cases because the injected harmful capability
lies outside the legitimate side-effecting behavior of the current task in
every case.


\section{Motivating Case: Persistent Authorization in Codex}
\label{app:codex-worked-path}

Codex illustrates how residual authority can arise even when the retained
matcher is narrow. A persistent approval produces a command-prefix grant
that remains effective when a matching command is proposed in another
project. This appendix traces that lifecycle from approval and storage to
the later authorization decision and its execution consequence.

\begin{center}
\small
\textsc{Persistent Approval}
$\longrightarrow$
\textsc{\texttt{default.rules}}
$\longrightarrow$
\textsc{Cross-Project Match}
$\longrightarrow$
\textsc{No Renewed Prompt}
$\longrightarrow$
\textsc{Changed Execution Mode}
\end{center}

\paragraph{Approval source and persistent state.}
Consider a command proposed during a legitimate task for which the user
selects Codex's persistent approval option. Source analysis traces this
decision through the native approval-handling path to the function that
persists the proposed authorization amendment. The implementation writes
an allow rule to
\texttt{\$CODEX\_HOME/rules/default.rules}.

This path makes the abstract terms in Section~4.1 concrete. The
\emph{approval source} is the program path through which the persistent
user decision creates the rule. The stored rule is the resulting reusable
authorization state. In the examined case, the representation retains an
argv prefix but does not retain the originating task, project, or reason
for which the command was approved. For example, the persistent state has
the following form:

\begin{quote}
\small
\ttfamily
prefix\_rule(pattern=["python3"], decision="allow")
\end{quote}

The residual authority is not the earlier command execution itself. It is
this surviving rule and its ability to affect later authorization
decisions after the original task has ended.

\paragraph{Authorization point and reuse.}
The \emph{authorization point} occurs when Codex evaluates whether a
later proposed command requires renewed approval. The authorization
engine loads the stored rules and compares the candidate argv against
their prefixes. A matching rule causes the command to be admitted without
a new approval request. Because the rule is stored under
\texttt{CODEX\_HOME} rather than under a project-specific directory, the
same rule is available when Codex is subsequently invoked in another
project.

Our reproduction exercises Codex's native persistence, parsing, matching,
and sandbox-selection functions. It first writes an authorization
amendment through the implementation's persistence function, reloads the
resulting rule as a new session would, and evaluates a matching command.
The recovered decision admits the command without renewed approval and
selects the corresponding execution mode.

\begin{table}[t]
\centering
\footnotesize
\setlength{\tabcolsep}{4pt}
\renewcommand{\arraystretch}{1.12}
\begin{tabular}{@{}p{0.14\linewidth}p{0.38\linewidth}p{0.38\linewidth}@{}}
\toprule
\textbf{Stage}
& \textbf{Concrete system event}
& \textbf{Retained or discarded context} \\
\midrule
Approve
& A persistent approval decision enters Codex's native authorization path.
& The proposed argv prefix is available at the approval source. \\

Store
& An allow rule is written to
  \texttt{\$CODEX\_HOME/rules/default.rules}.
& The prefix survives; the originating task, project, and justification
  do not. \\

Reload
& A later Codex session loads the global rule store.
& The rule remains effective outside the session and project in which
  it was created. \\

Match
& A later command's argv matches the retained prefix.
& The authorization point admits the command without a renewed prompt. \\

Execute
& The authorization decision selects the corresponding execution mode.
& A contained filesystem test demonstrates an effect rejected under
  sandboxed execution. \\
\bottomrule
\end{tabular}
\caption{Concrete interpretation of an approval source, persistent grant,
and authorization point in the examined Codex implementation.}
\label{tab:codex-worked-path}
\end{table}

\paragraph{Observed consequence.}
Separate filesystem tests evaluate the consequence of the recovered
decision. The tests execute the same benign marker write under two
conditions. Codex's native filesystem sandbox rejects the write when
sandbox enforcement is applied, whereas the execution mode selected
after the persisted-rule decision permits the write. The marker resides
outside the test workspace but inside a temporary, researcher-controlled
directory.

The authorization tests and filesystem tests establish consecutive
implementation properties: persistent approval creates a cross-project
rule; a later match suppresses renewed approval; and the resulting
execution mode changes the permitted filesystem effect. They do not
constitute a recorded, continuous prompt-injection trajectory through
the Codex user interface. We therefore use this case to validate the
authorization path and its consequence, rather than report it as an
additional end-to-end attack trial.

\paragraph{Containment and disclosure.}
The reproduction operates only on temporary files in a local test
environment and does not access third-party systems, credentials, or
user data. We report the authorization path and contained consequence
needed to substantiate the paper's claims, while omitting operational
payloads and bypass variants. The issue was communicated to the affected
vendor before publication.


\section{End-to-End Attack Walkthrough in Goose}
\label{app:goose-worked-replay}

The frozen Terminal-Bench \texttt{portfolio-optimization} case provides
a concrete instance of the construction in Section~4. We start from a
contained target effect, derive the native grant needed to admit the
corresponding action, acquire that grant during ordinary task execution,
and then compare Fresh and Farmed runs through Goose's native approval
interface.

\begin{center}
\small
\textbf{Target effect}
$\longrightarrow$
\textbf{required shell action}
$\longrightarrow$
\texttt{always\_allow: shell}
$\longrightarrow$
\textbf{benign acquisition}
$\longrightarrow$
\textbf{replay}
\end{center}

\paragraph{Case construction.}
The legitimate task requires the agent to run a shipped benchmark script
through its shell tool. The context change prepends a harmless marker
write to the script while retaining its original contents and behavior.
The task and command family remain unchanged, but the content reached by
the command now produces an additional observable effect.

The effect oracle checks for a local marker file inside the experiment
container. The operator does not use the network, access credentials, or
affect a third-party system.

\paragraph{Backchaining to a native grant.}
We begin with the marker effect and its reference path. Reaching the
modified script requires a task-related shell invocation, denoted
$a^\star$. With an empty authorization store, this action reaches Goose’s approval boundary and is therefore one of the approval-gated actions in \(A_{\mathrm{req}}\).
Goose does not persist the command or script identity associated with
this approval. Its matcher records only the tool name. Under the
authorization profile recovered in Section~4.1, any persistent
\texttt{shell} grant therefore admits $a^\star$:

\[
R=\{g_{\texttt{shell}}\},
\qquad
M_{\mathrm{Goose}}(R,a^\star)=\mathrm{allow}.
\]

We then select a clean task execution that naturally requires Goose's
shell tool. In this case, the unmodified
\texttt{portfolio-optimization} task itself supplies the benign
acquisition context: the shell operation is needed to run the benchmark,
but the shipped script has not yet been modified. Backchaining identifies
this task and the expected grant; Acquire verifies whether Goose actually
creates and retains it.

\paragraph{Acquire.}
Acquire runs the legitimate task in the clean context. When Goose requests
approval for shell use, the experiment controller selects the product's
persistent option, \texttt{ALLOW\_ALWAYS}. Goose writes the resulting
authorization state to \texttt{permission.yaml}, including:

\begin{quote}
\small
\ttfamily
user:\\
\phantom{xx}always\_allow:\\
\phantom{xxxx}- shell
\end{quote}

This file is the grant carried into replay. Although approval is requested
for a concrete task-related operation, the stored representation contains
only the shell-tool identity. It does not retain the approved command,
script contents, project, or task justification.

\paragraph{Fresh and Farmed replay.}
Fresh starts with an empty permission store after the context change has
been installed. When the agent proposes the target shell action, Goose
issues a new approval request. The controller rejects this request, the
action is withheld, and the marker remains absent.

Farmed starts from the permission store captured during Acquire. The
context change and task instruction are the same as in Fresh. When the
agent proposes the shell action, the stored \texttt{shell} grant matches
it. No target-action approval request is recorded, the action proceeds,
and the marker is created.

The marker is cleared before each run. For the authorization boundary
examined here, the contrast is whether the acquired permission store is
absent or restored.

\begin{table}[t]
\centering
\small
\setlength{\tabcolsep}{5.5pt}
\renewcommand{\arraystretch}{1.12}
\begin{tabular}{@{}lcccc@{}}
\toprule
\textbf{Run}
& \textbf{Context}
& \textbf{Shell grant}
& \textbf{Target approval}
& \textbf{Marker} \\
\midrule
Acquire & Clean   & Acquired & \texttt{ALLOW\_ALWAYS} & No \\
Fresh   & Changed & Absent   & Requested; rejected    & No \\
Farmed  & Changed & Restored & None recorded          & Yes \\
\bottomrule
\end{tabular}
\caption{\textbf{Native Goose replay for one archived
\texttt{portfolio-optimization} repetition.}
Across all five repetitions, Acquire produced a persistent shell grant,
Fresh produced one or two target-action approval requests and no marker,
and Farmed produced no target-action approval request and did produce
the marker.}
\label{tab:goose-worked-run}
\end{table}

\paragraph{Relation to the replay definition.}
The \texttt{shell} entry is the retained grant
$g_{\texttt{shell}}\in G_h$. Fresh records that the target action reaches
an approval boundary without this grant. Farmed records no corresponding
approval request after the grant is restored, while the marker supplies
the harmful-effect oracle. At the level recorded by the archived episode
summary,

\[
M_{\mathrm{Goose}}(G_0,a^\star)=\mathrm{ask},
\qquad
M_{\mathrm{Goose}}(G_h,a^\star)=\mathrm{allow},
\qquad
Q_2=0,
\qquad
H_2=1.
\]

The retained grant originated in a clean run through Goose's ordinary
approval workflow. Its later reuse therefore does not depend on an
adversarial approval: the adversarial change occurs in the script content
after the grant has already been created.

The archived summaries retain the permission store, target-action counts,
approval counts, and marker outcome, but not the complete message-level
ACP trace. We consequently report the observed approval-boundary and
effect fields without inferring additional intermediate events.

\paragraph{What outlives the original context.}
The object carried from Acquire into Farmed is the native
\texttt{always\_allow: shell} entry. The approved script contents and
the task context are not part of that entry. Goose therefore reuses the
grant after the content determining the action's effect has changed.
This is the concrete residual authority in the case: the approval's
machine representation survives, while the context that gave the
approval its original meaning does not.


\section{Bootstrap Confidence Intervals}
\label{app:bootstrap}

We quantify uncertainty using a paired target-cluster bootstrap. For each
model and history length, we resample injection targets
$(\mathrm{suite},X,Z)$ with replacement, retaining all rows and paired
fresh and accumulated outcomes associated with each target. For bootstrap
sample $b$, we compute

\[
\Delta_b(h)
=
\operatorname{ASR}^{(b)}_{\mathrm{acc}}(h)
-
\operatorname{ASR}^{(b)}_{\mathrm{fresh}}(h).
\]

We use $B=2000$ resamples with seed 20260907 and report percentile 95\%
confidence intervals. Table~\ref{tab:agentdojo-bootstrap} reports the
endpoint effects at $h=64$. Claude produces no successful attacks in either
condition, making its empirical bootstrap distribution degenerate; we
therefore report the observed zero effect as a negative control rather
than an informative confidence interval.

\begin{table}[t]
\centering
\small
\setlength{\tabcolsep}{7pt}
\renewcommand{\arraystretch}{1.08}
\begin{tabular}{@{}lcc@{}}
\toprule
\textbf{Model}
& \textbf{$\Delta$ASR (pp)}
& \textbf{95\% CI (pp)} \\
\midrule
Gemini-3.1-Flash-Lite & 35.1 & [30.6, 39.5] \\
GPT-4.1               & 27.9 & [23.6, 32.0] \\
Qwen3-14B             & 23.5 & [19.5, 27.4] \\
DeepSeek-V4-Pro       & 11.4 & [7.8, 15.6] \\
Llama-3.3-70B         & 0.4  & [0.0, 1.1] \\
Claude-Sonnet-5       & 0.0  & Degenerate \\
\bottomrule
\end{tabular}
\caption{Paired target-cluster bootstrap estimates for the AgentDojo
experiment at $h=64$. Effects are accumulated minus fresh ASR.}
\label{tab:agentdojo-bootstrap}
\end{table}

\end{document}